\documentclass[conference]{IEEEtran}
\IEEEoverridecommandlockouts
\usepackage{cite}
\usepackage{amsmath,amssymb,amsfonts}
\usepackage{algorithmic}
\usepackage{multicol}%
\usepackage{graphicx}
\usepackage{textcomp}
\usepackage{siunitx}
\usepackage{booktabs}
\usepackage{fvextra}
\usepackage{array} %
\usepackage{float}
\usepackage{url}
\usepackage{enumitem} %
\usepackage{caption}
\floatstyle{plaintop}
\restylefloat{table}

\usepackage{xcolor}
\def\BibTeX{{\rm B\kern-.05em{\sc i\kern-.025em b}\kern-.08em
    T\kern-.1667em\lower.7ex\hbox{E}\kern-.125emX}}

\usepackage{lipsum}%
\usepackage{tcolorbox}

\newfloat{codelisting}{h}{lop}
\floatname{codelisting}{Listing}

\renewenvironment{codelisting}
  {\captionsetup{type=codelisting}}
  {}

\tcbuselibrary{breakable}
\tcbset{%
  width=0.5\textwidth,
  halign=justify,
  center,
  breakable,
  colback=white    
}

\definecolor{dataEntry}{HTML}{f16667}
\definecolor{click}{HTML}{57c7e3}
\definecolor{APICall}{HTML}{8dcc93}
\definecolor{Location}{HTML}{4c8eda}
\definecolor{DOMEffect}{HTML}{ecb5c9}
\definecolor{annotatedVertex}{HTML}{a6c9ec}

\renewcommand{\paragraph}[1]{\textbf{#1 }}

\begin{document}

\title{Token Efficient Task Execution via Application Behavior Modeling for Web Agents

}

\author{\IEEEauthorblockN{Alexandru Ianta}
\IEEEauthorblockA{\textit{Department of Computing Science} \\
\textit{University of Alberta}\\
Edmonton, Canada \\
0000-0003-0276-0620}
\and
\IEEEauthorblockN{Eleni Stroulia}
\IEEEauthorblockA{\textit{Department of Computing Science} \\
\textit{University of Alberta}\\
Edmonton, Canada \\
0000-0002-8784-8236}
}

\maketitle

\begin{abstract}
The strong performance of AI Agents across an impressive variety of tasks is driving an unprecedented investment in agentic infrastructures, however the cost of processing tokens is fast increasing. Web agents automate the execution of web-application tasks described in natural language, by analyzing the web-application's user interface (UI) and interacting with it. This work introduces OdoBot, a novel web-agent architecture that completes tasks at a fraction of the cost when compared to conventional web agents. This is achieved by leveraging a behavioral model of the underlying application constructed by analyzing successful task-execution demonstrations. 
Our experiments with 45 tasks on the Canvas Learning Management System (LMS) demonstrate that OdoBot uses 44\% and 80\% fewer tokens than two state-of-the-art competitor agents (Agent-E and WebVoyager), while also surpassing WebVoyager in terms of task success rate.

\end{abstract}


\section{Introduction}

As academia and industry explore how to leverage AI agents to replicate human skills and automate tasks, the automation of user interactions with web applications has emerged as a prevalent test bed for web agents. In this context, web agents are given a task description in natural language and interact independently with the web application to complete it. These agents employ Large Language Models (LLMs) to interpret the current web-application state, break the task down into simpler steps, and execute them through the application's user interface(UI)~\cite{survey-web-agents, LM-agent-survey, xi2023risepotentiallargelanguage}. As LLMs improve, so do web agents~\cite{zhou2026externalizationllmagentsunified}. However, tokens, the LLMs' data-processing units, are becoming ever more expensive~\cite{cottier2025risingcoststrainingfrontier, economist_scrabling_to_curtail}, and the web-agents that rely on them can become unaffordable for the organizations that rely on them to support their users.

Web agents treat web applications as environments which they can observe and act upon, in pursuit of executing some task~\cite{survey-web-agents, LM-agent-survey}. 
Contemporary web applications are typically implemented as a stack of a back-end server managing the application data, a front-end layer responsible for the data exchange between the user and the back-end, and a user interface (UI) defined by the Document Object Model (DOM) that the front-end layer serves to the user to support the user interaction.
Contemporary web agents observe the application state through its UI, capturing observations textually (through DOM~\cite{zheng2024synapse, lutz2024wilbur, abuelsaad2024agenteautonomouswebnavigation} or Accessibility Tree~\cite{he2024webvoyager} snapshots), or visually (through detailed screenshots~\cite{he2024webvoyager, see-act}). The cost of the LLM usage for analyzing these observations is analogous to the size and complexity of the application UI and the accuracy of the ``observation lens''. As well, newer more powerful LLMs tend to require more and more expensive tokens for analyzing the same observation, driving the cost of web-agent deployment further up.

In this work, we introduce OdoBot, a web agent that puts forward a novel observation methodology, based on the construction of a behavioral model of the underlying application. In addition to conventional DOM snapshot observations of the UI, OdoBot monitors the browser's location (the url in the address bar), and the network communication between the front-end and the back-end. This enables OdoBot to track its progress during task execution, by deterministically comparing live browser location changes and network communication to previous observations, as opposed to sending the execution history to an LLM. In addition, OdoBot analyzes previously observed DOM snapshots to detect application-specific common HTML sub-structures that identify regions of the screen containing related elements. During task execution, only regions of the DOM relevant to the action being executed are sent to the LLM for processing. In these ways, OdoBot significantly reduces its token consumption during task execution in exchange for an upfront investment for model construction, fast amortized with a relatively small number of executed tasks.   

To evaluate our work, we compare OdoBot, against WebVoyager~\cite{he2024webvoyager} and Agent-E~\cite{abuelsaad2024agenteautonomouswebnavigation}, visual and textual state-of-the-art agents respectively, by having each agent complete 45 tasks from a novel realistic evaluation environment built on the Canvas Learning Management System (LMS). To our knowledge, while there are many works investigating the use of generative AI in teaching contexts~\cite{educsci14060636genaiteachingsurvey}, this is the first evaluation environment for web agents in this domain.
 
The rest of this paper is organized as follows. Section \ref{sec:background} places our work in the context of recent literature. Section \ref{sec:methodology} describes our model-construction methodology, and  Section \ref{sec:task-execution} describes OdoBot's model-driven task execution process. Section \ref{sec:sandbox} describes the construction of the Canvas-based sandbox. Sections \ref{sec:study} and \ref{sec:findings} report on the design of our comparative-evaluation study and our findings. Finally, Section \ref{sec:conclusions} concludes with a statement of the contributions of this work and outlines our plans for future work.

\section{Background and Related Research}
\label{sec:background}

\paragraph{Web-Agent Test Beds}
The performance metric typically used to evaluate web-agents' performance is ``task success rate (SR)", i.e., the percentage of tasks from an evaluation task set they can successfully complete. To date, several test beds have been developed.

Mind2Web is constructed from crowd-sourced trajectories captured from 137 real-world websites. It is an offline dataset, which means that web agents evaluated on Mind2Web must execute tasks in the exact same way as captured in the collected trajectories; any deviation is counted as a failure, leading to false negatives~\cite{NEURIPS2023_mind2web}. This is especially problematic when evaluating task success on web applications where often times there are multiple valid ways to achieve the same objective. 

WorkArena, and WebArena offer online environments for the evaluation of web agents~\cite{drouin2024workarenacapablewebagents, ICLR2024_4410c071web_arena}, where agents can interact with live hosted environments. 
WorkArena is built around the ServiceNow platform, where users can construct and manage their own applications~\cite{service-now}. The WorkArena test applications are constructed using the ServiceNow software framework, and as a result they are not necessarily representative of contemporary web applications.
WebArena offers a collection of applications purposefully developed to mimic real-world counter parts in the e-commerce, social forum, collaborative software development, and content management system domains~\cite{ICLR2024_4410c071web_arena}. 

The WebVoyager benchmark is similar to Mind2Web, consisting of 643 tasks over 15 websites~\cite{he2024webvoyager}. However, unlike Mind2Web, the WebVoyager benchmark does not contain trajectories for all these tasks, and unlike WorkArena and WebArena evaluation is not done on hosted environments but instead on the open web. Though more realistic, the open web environment does risk tasks becoming impossible as third party content changes. Evaluation of task success on the WebVoyager benchmark is done manually by humans, or at scale using an LLM evaluator, by analyzing an agent's trajectory and determining whether or not the trajectory satisfies the objectives of the task. 

In this paper, we contribute to the currently available realistic environments by offering an evaluation dataset for web agents on Canvas, a popular Learning Management Systems (LMS).  Such an environment could help evaluate web agents interacting with the same platforms as students. Additionally, the dataset on which OdoBot is evaluated focuses on tasks that make material changes to the application state via network requests. In doing so, it is possible to use the provided evaluation scripts to detect the presence of the network requests that indicate task success in a deterministic fashion, instead of relying on LLM verification.

\paragraph{Web Agents}
Web agents break down tasks into smaller steps that can be carried out by applying actions to the UI. The execution plan for the task is updated to reflect the action, a new observation of the UI is made, and the process repeats~\cite{LM-agent-survey, xi2023risepotentiallargelanguage, survey-web-agents}. Sequences of observation-action pairs form \textit{trajectories}~\cite{LM-agent-survey, survey-web-agents, zheng2024synapse}.
Focusing on the readily observable UI is a reasonable starting point, and allows for a clean one-to-one association between actions and observations. However the UI represents only a portion of an application's `true' state, by re-framing trajectories as sequences of events, some of which are associated with (user invoked) actions, while others are emitted by the application itself (network events), OdoBot can make use of an expanded observation space in comparison with previous work.    

Most agents keep the trajectory of their current task execution in their LLM context window to keep themselves `aware' of what they have already done, and what they have yet to do~\cite{survey-web-agents, xi2023risepotentiallargelanguage, he2024webvoyager}. WILBUR~\cite{lutz2024wilbur} and  NNetNav~\cite{murty2025nnetnavunsupervisedlearningbrowser} go a step further and process their own trajectories into `demonstrations' for future retrieval in an effort to learn from past experiences. However, OdoBot, is the first agent to drive task execution through a behavioral model of the application constructed using past trajectories. NNetNav achieves a 35.2\% task success rate using Llama-8B, while WILBUR achieves a 52.6\% task success rate using OpenAI's GPT-4 on the WebVoyager benchmark~\cite{lutz2024wilbur}. 

Agents like WebVoyager can perceive the UI through both textual representations (accessibility tree snapshots) or visually, through screenshots~\cite{he2024webvoyager}. WebVoyager achieves a 44.3\% and 59.1\% task success rate (with its text-based and visual-based configurations respectively) on its test bed of 643 task instances across 15 popular websites, relying on OpenAI's closed source GPT-4 LLM. 

OdoBot, Synapse~\cite{zheng2024synapse}, WILBUR~\cite{lutz2024wilbur}, and Agent-E~\cite{abuelsaad2024agenteautonomouswebnavigation} are strictly textual agents. For textual agents, full snapshots of the DOM are `noisy and expansive' often exceeding an LLM's maximum allowable input (context window)~\cite{abuelsaad2024agenteautonomouswebnavigation}. This makes intelligent refinement of the DOM critical.  Zheng et al. used minimum-spanning trees of likely relevant elements in the DOM to improve task success rate from 0.4\% to 3.2\% on the Mind2Web benchmark~\cite{zheng2024synapse}. WILBUR's DOM observations are lists of leaf elements only including their inner text, tag type, and a set of curated HTML attributes. . Agent-E uses; the inner text of the `body' element, a JSON representation of specific input-related elements, or a JSON representation of all elements, depending on whether it is looking for information, entering data, or listing all inter-actable elements on the page, respectively~\cite{abuelsaad2024agenteautonomouswebnavigation}. Agent-E achieves a state-of-the-art 73.1\% task success rate on the WebVoyager benchmark.

\begin{table}[]
\begin{tabular}{|m{2cm}|m{1.6cm}|c|m{2.8cm}|}
\hline
\textbf{Agent}  & \textbf{Task SR} & \textbf{Year} & \textbf{LLM} \\ \hline
Agent-E & 73.1\% \cite{abuelsaad2024agenteautonomouswebnavigation} & 2024 &  gpt-4-turbo* \\ \hline
WebVoyager & 59.1\% \cite{he2024webvoyager}  & 2024 &  gpt-4-1106-preview                              \\ \hline
WILBUR                                            & 52.6\% \cite{lutz2024wilbur} & 2024 &  gpt-4-0125-preview                                      \\ \hline
WebVoyager {[}Text-only{]}               & 44.3\% \cite{he2024webvoyager} & 2024 &  gpt-4-1106-preview                                       \\ \hline
NNetNav                                           & 35.2\% \cite{murty2025nnetnavunsupervisedlearningbrowser}  & 2025 &  Llama8B                                       \\ \hline
\end{tabular}
\caption{Reported task success rate (SR) on the WebVoyager benchmark for different agents. *The exact snapshot for Agent E is not reported~\cite{abuelsaad2024agenteautonomouswebnavigation}. }
\label{tab:event-node-index}
\end{table}

\section{Model Construction}
\label{sec:methodology}

OdoBot treats web applications as emitters of events. Events indicate changes in some aspect of the application's state. They can be triggered by external actions applied by a user or web agent, like a click, or by internal processes of the application, like the front-end transmitting a user's credentials to the backend for authentication. A web application can be characterized by the possible sequences of events it can produce. A trajectory is any sequence of events. By capturing trajectories corresponding with the execution of meaningful tasks on an application, it is possible to create a behavioral model that organizes the expected sequences of events associated with the completion of those tasks. Thus the problem of task execution on web applications becomes one of re-creating an observed sequence of events corresponding with the execution of the task by navigating the behavioral model. The construction of this application behavioral model is a three-step process, involving (i) trajectory collection, (ii) graph model construction, and (iii) graph annotation.

\subsection{Trajectory Collection}
\label{sec:trajectory-collection}

OdoBot defines \textit{tasks} in terms of high-level natural language descriptions of some operation(s) or outcome(s) on the underlying web-application that result in material server-side changes, i.e., side-effects. For example, \emph{creating a new post} is a task for a blog platform. \textit{Task instances} are descriptions of tasks at a sufficient level of detail to permit their execution. For example, an instance of the above task could be \emph{create a new post, with the title ``Hello world'' and ``This is my first post!'' as its content.}. 

To construct a model that supports the tasks possible on the subject web application, OdoBot collects \textit{trajectories}, i.e., a chronologically ordered sets of\textit{events}. OdoBot employs, OdoX, a browser extension 
to monitor user-application interactions and record the trajectories. 

There are ten different types of trajectory events, organized in two categories. \textit{User interaction events} include click, data entry, tinymce, radio button, select, and checkbox events. \textit{Application events} include non-GET REST API calls, graphQL mutation operations, location events and DOM effects. The triggers, key properties and some examples of these properties are listed in table~\ref{tab:event-table} for each event type. 

\begin{table*}[h!]
    \centering
    \includegraphics[width=\linewidth]{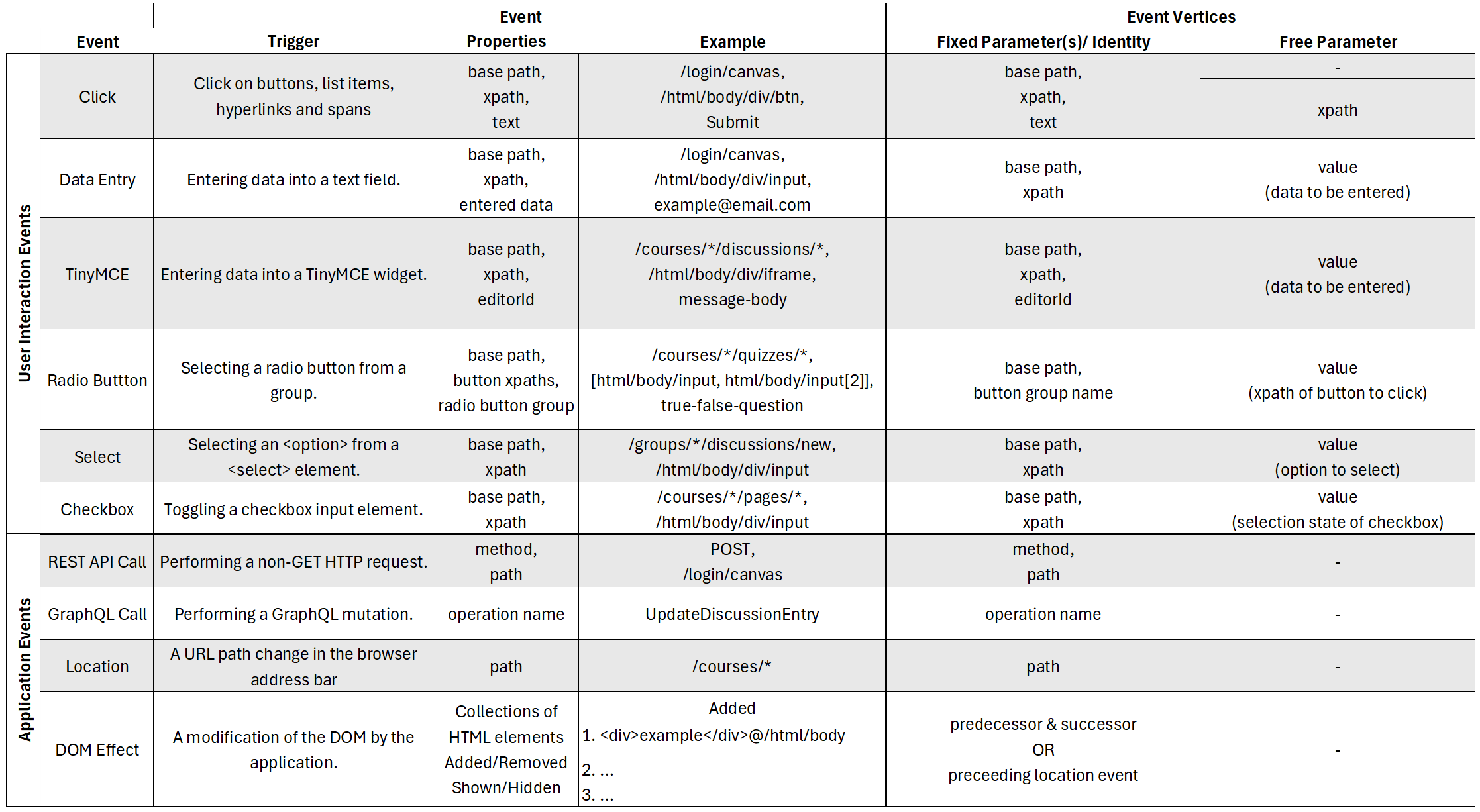}
    \caption{Trajectory events (left) and their mapping to event vertices (right). Comparing the `properties' and `fixed parameter' columns shows that for two click events to map to the same event vertex they must share base paths, xpaths and text values.   
    }
    \label{tab:event-table}
\end{table*}

\paragraph{User Interaction Events} are triggered by the user and represent data-entry activities, through a variety of UI widgets. 
For these events, OdoBot records an HTML snapshot of the Document Object Model (DOM), the base path of the URL at capture time, the HTML element with which the user interacts and its key properties, as shown in Table~\ref{tab:event-table}. 
The HTML snapshots are filtered to remove scripts, embedded CSS styles, and the contents of SVG paths. 
The base path of an event is the URL at which it was captured minus any protocol, host and port information. For example, a user interaction event captured at `http://localhost:8088/course/2/assignment/7/submission' has a base path of `/course/*/assignment/*/submission'. `*' wildcards are used to replace dynamic portions of the URL. This was achieved by manually constructing a regex that captured dynamic URL segments as defined in the underlying application's API documentation.

Button clicks, data-entry boxes, selection menus, and checkboxes are associated with xpath(s) from the document root to their corresponding HTML elements. Radio button events include xpaths to all the options in the same group. 
For example, the xpath `/html/body/div[2]/span' refers to the span element located in the second div of the body element of the HTML document. These xpaths allow the retrieval of the related HTML elements from the event snapshot. This allows OdoBot to capture additional event-specific information, like any text associated with buttons, placeholder texts in input fields, what options are available in a radio group or select drop-down box, etc.

TinyMCE events are a special case because they are associated with rich content editors embedded into the page via iframes whose inner contents are not addressable through xpaths. Therefore, in addition to an xpath to the iframe, TinyMCE events record an `editor id' given by TinyMCE, which can be used to interact with the editor instance via Javascript.  

\paragraph{Application Events} are triggered in response to user interaction events, and they include (i) network operations that transmit user data to the back end, such as non-GET REST API calls and graphQL mutation operations, and (ii) changes in the state of the application's DOM and UI.
Only non-GET API calls and mutation operations are of interest because they correspond with material changes on the server-side of the application, and therefore possible task objectives. 
UI state changes include changes to the URL in the browser's address bar (location events) corresponding with the transition between pages or views, or the addition, removal, hiding, or revealing of HTML elements in the DOM (DOM effects). DOM effect events organize their elements into lists corresponding with DOM additions, removals, etc. This is the only application event type that also includes filtered HTML snapshots of the DOM that are otherwise typical for user interaction events. Note that, like base paths, paths captured for location and API call events are also normalized (dynamic portions of the paths are replaced by an `*'). 

\paragraph{Semantic Event and Trajectory Descriptions}
The trajectories that OdoBot uses to construct the application graph model are demonstrations of the tasks that the application supports. OdoBot uses an LLM to generate natural language descriptions for every trajectory event and then summarizes these event descriptions into a trajectory description. Listing~\ref{lst:click-event-description} shows an example description generated for a click event in a trajectory. 

\begin{codelisting}
\vspace{.2cm}\hrule
  \caption{\label{lst:click-event-description}The description of a click event produced by an LLM during the trajectory description synthesis process.}
  \begin{Verbatim}[breaklines=true]
The user clicked the "Scenario Builders" link in the Groups tray (anchor href="/groups/10"), selecting that specific group. This click initiated navigation to the group's page (/groups/10) from the dashboard.
\end{Verbatim}
\hrule\vspace{.2cm}
\end{codelisting}

The prompt used to generate the trajectory-level description explicitly instructs the LLM to write in the imperative style and avoid including task-specific inputs, focusing instead on high-level descriptions of application features. This aligns trajectory-level descriptions with task definitions and enables OdoBot to map new task requests to trajectories it has observed. Listing~\ref{lst:trajectory-description-sample} shows a generated trajectory-level task description. 

\begin{codelisting}
\vspace{.2cm}\hrule
  \caption{\label{lst:trajectory-description-sample}The description of a trajectory produced by an LLM.}
  \begin{Verbatim}[breaklines=true]
Log in to your account. Within a group's Pages/wiki area, create a new wiki page by providing a title and content, enable the option to add the page to students' to‑do/notify them, and save/publish the page so the app opens the newly created page view.
\end{Verbatim}
\hrule\vspace{.2cm}
\end{codelisting}

The trajectory-level task descriptions generated in this step are subsequently converted into vector embeddings using OpenAI's `text-embedding-3-large' model and stored in a vector database.

\subsection{Graph Model Construction}
\label{sec:graph-construction}

OdoBot constructs a graph model from a collection of trajectories. The graph vertices represent a collection of equivalent events that OdoBot has observed in the collected trajectories. Edges correspond to the temporal events order, so if an event is mapped to vertex A and the subsequent event in the trajectory is mapped to vertex B, an edge is inserted from vertex A to vertex B. 

Two events are equivalent, if they have the same fixed parameters. Table~\ref{tab:event-table} shows how this equivalence is computed for each event type. For example, the fixed parameters of click events are base path, xpath, and text properties. Therefore, if two click events have the same base path, xpath and text values, they are equivalent and get mapped to the same vertex. 

To construct the graph, OdoBot iterates through all events in all trajectories. For each new event, OdoBot queries the graph for a corresponding event vertex using the event's fixed parameter values. For example, for a click event captured at the base path `/login/canvas' with the xpath `/html/body/div[3]/div[2]/input', OdoBot searches for a click event vertex with matching base path and xpath values. For a REST API call, OdoBot searches for a API call event vertex with matching HTTP method and path values. 
If no matching vertex is found, a new vertex is created with a unique randomly generated ID and the event's fixed parameter values. In this way, every event corresponding to `clicking the login button', irrespective of the specific trajectory in which it belongs, will be mapped to the same event vertex in the graph; every event corresponding with entering some text into a particular input box will map to the same event vertex in the graph; and so on. 

DOM Effect events are a special case because they contain four (added/removed/shown/hidden) lists of elements whose contents and element order vary due to stochastic factors, such as network latency. As a result, the DOM effect events after a page load may look slightly different from each other, making it difficult to use any single set of properties in the underlying elements as fixed parameters to identify the DOM effect. This is why the equivalence of DOM Effect events is defined by their preceding and succeeding events: DOM Effect events that happen after some event and before a subsequent event belong to the same DOM Effect event, unless the preceding event is a location event, in that case any subsequent DOM effect is grouped together in a single vertex after the location event. This carve out for location events exists because all DOM effects following a location change represent the loading of elements onto the page. In other contexts, the successor event still provides meaningful information, for example: entering some invalid input might cause DOM changes highlighting the error (eg: `username is already taken'), the subsequent data entry on the same input box, emitted when the user changes the value to fix the error, separates those DOM effects from DOM effects informing the user that the data validates correctly (eg: `username is available'), which might instead be followed by a click on the submit button.

In practice, this means that graph construction happens in three passes through all the trajectories. First, all events except DOM Effects are processed into event vertices. Next, DOM effects are processed and result in the generation of new event vertices between vertices that were generated in the first pass. Finally, directed edges are added connecting everything together following the temporal order observed in the trajectories. This ensures that corresponding events for all possible predecessors and successors to DOM effect events are defined when they are needed to create DOM effect events. 

\begin{figure*}[h!]
    \centering
    \includegraphics[width=\linewidth]{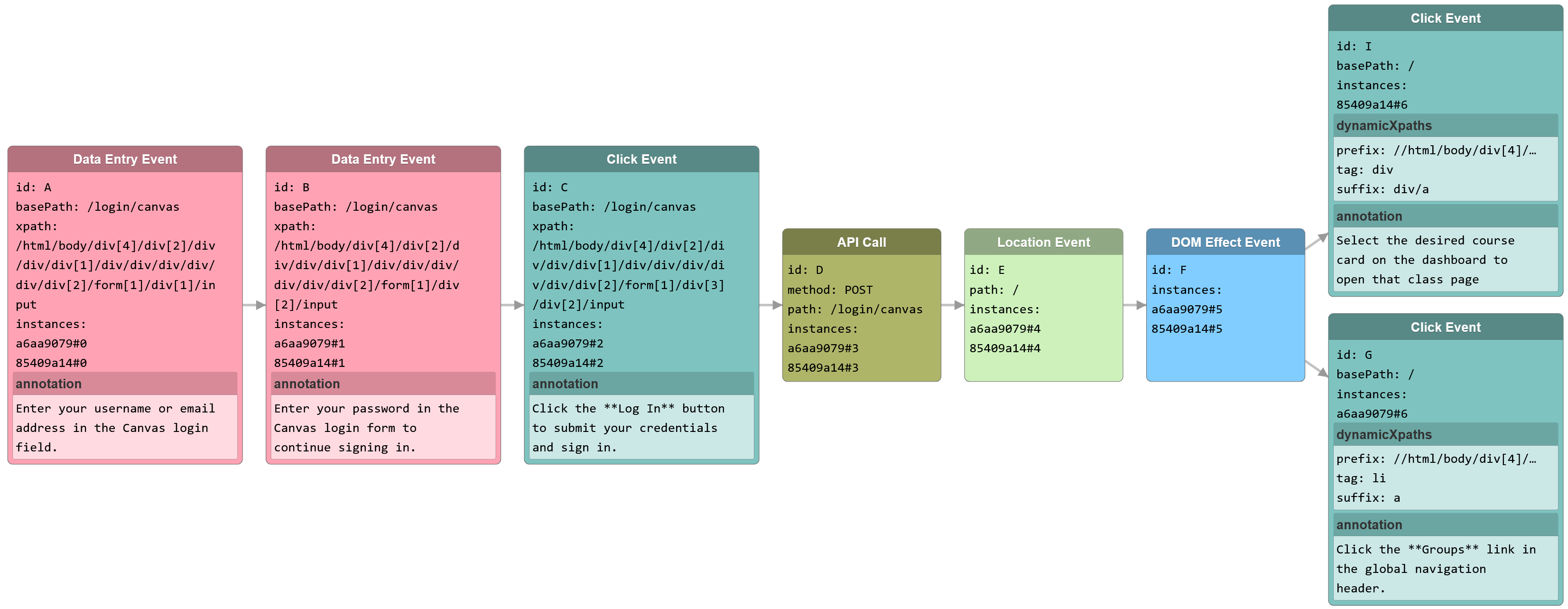}
    \caption{A sample graph model, generated from trajectories a6aa9079 and 85409a14, that diverge only in the last user action.}
    \label{fig:graph-model-sample}
    \vspace{-0.5cm}
\end{figure*}

Each event vertex in the model has an `instances' property that contains the ID of every event that was mapped to it. The size of this set corresponds to how common a particular event is among the trajectories used to build the model. Event IDs combine a unique ID for the trajectory and the index of the event using a `\#'. Figure~\ref{fig:graph-model-sample} shows a sample model with these instances mapped to events. 

In addition to their fixed parameters that are used to define equivalence, events also have free parameters whose values must be resolved at task execution time. This distinction between fixed and free parameters enables OdoBot to generalize its observations, and execute new task instances based on similar task instances captured in the trajectories used to construct the model. For example, from table~\ref{tab:event-table}, the location (base path and xpath) of a text input field are fixed parameters of data entry events, while the value to be entered is a free parameter, meant to be resolved at task execution time with a task appropriate value.

Figure~\ref{fig:graph-model-sample} shows the graph constructed by analyzing two trajectories (a6aa9079 and 85409a14). In one trajectory, the user logged in with their username and password and selected a course card from the dashboard of the application. In the other trajectory, the user logged in with a username and password and clicked on the groups link in the navigation bar.
The first two nodes (in pink) correspond to two data entry events, as the user enters their username and password. The turquoise node corresponds to the login button click, followed by the API call application event, submitting the login credentials to the back end. In response, the application location changes to the dashboard (in green), and a set of DOM effects occur (blue) as the dashboard elements are loaded. Finally, the user chooses to click on a course card (in turquoise) or a click on an item in a navigation bar.

\subsection{Graph Annotation}
\label{sec:model-annotation}

Once the application model graph has been constructed, OdoBot proceeds to annotate it with information about (i) click events whose parameters can only be determined at execution time, and (ii) the semantics of the graph vertices. In each of these cases the annotation process results in the addition of a new property to relevant vertices in the graph shown as the `annotation' and `dynamicXpath' properties in Figure~\ref{fig:graph-model-sample}.

\subsubsection{Free Parameter Click Events}

At this stage, all click events in the graph are modeled as having fixed xpath parameters. That is, every click-event vertex refers to the elements that were clicked in the observed trajectory clicks. An example of a fixed parameter click event would be clicking on the login button. At task execution time, the same login button that was observed in the trajectories can be used to login again. This is not, however, the case with all click events; sometimes, the element that needs to be clicked is specific to the task at hand, such as, a link to a particular post to be edited, a specific element from a menu bar, a card corresponding to a particular course on a dashboard. These are click events whose free parameter should be the xpath of the element to be clicked. That is, the element which must be clicked has to be determined at task execution time, from a set of options that can also only be determined at task execution time.

Ideally, the HTML element containing the free-parameter options is an ordered (`ol') or unordered (`ul') list element, and the options are (`li') child elements of that list. Click events with target element xpaths leading inside an `ol' or `ul' element could then be used to identify a click event whose parameter ought to be free and resolved to the correct element at task execution time. However this sort of structure is rarely enforced, and indeed many web applications construct lists of structurally complex children using div, span, and other general tags.

To robustly identify elements that act as containers of related options, OdoBot 
relies on the assumption that the web application adheres to web-design best practices dictating that visually similar elements should behave similarly and should be grouped together~\cite{ux-similarity-1}. Thus, by mining common HTML sub-structures from the DOM snapshots, OdoBot can find sets of objects that are related. If those objects share a common HTML parent element, that parent element is assumed to be the container of those objects. 

To produce a database of common HTML sub-structures, DOM snapshots are converted into graphs (trees), where structural information around every vertex of those trees is hashed using the weisfeiler-lehman subgraph hashing technique~\cite{wl-shingles-theory}. The resulting hashes are used to produce a min-hash~\cite{mining_massive_datasets} of the local HTML structure around every node in the DOM. Local sensitivity hashing (LSH)~\cite{mining_massive_datasets} is used to estimate the distances between min-hashes, and DBSCAN~\cite{sklearnDBSCAN} uses the resulting distance metric to identify clusters of similarly structured elements in the snapshot. When elements in a cluster share the same parent element in the DOM, a set of common HTML sub-structures and their container is identified, and their xpaths are saved. OdoBot checks if the xpath selects an element inside any of the identified common HTML sub-structure containers. If so the click event associated with that event is annotated as having a free parameter to be resolved at task execution time.

\begin{figure}[h!]
    \centering
    \includegraphics[width=\linewidth]{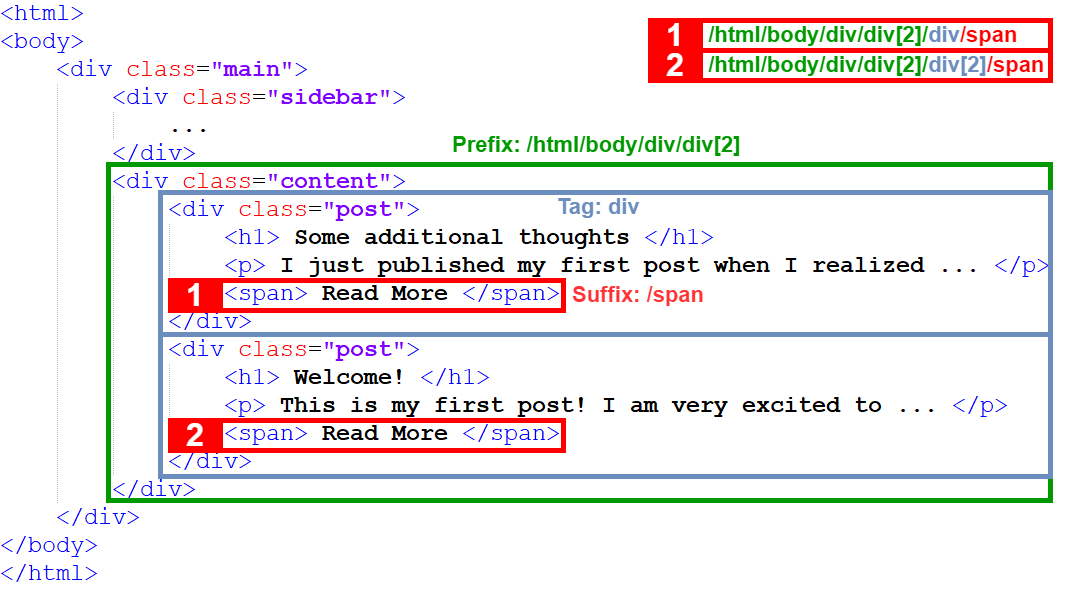}
    \caption{An illustrated example of a (prefix, tag, suffix)-tuple dynamicXpath. The green rectangle highlights the prefix container element. The blue rectangles highlight repeated HTML sub-structures identified by matching children with the dynamicXpath's tag. The red rectangles highlight the elements matching the suffix. A click event annotated with this dynamicXpath can be resolved by clicking either element 1 or 2, and a click event whose xpath is either that of element 1 or 2 will be identified as appertaining to that click event. The color-coded element xpaths are shown in the top right. }
    \label{fig:illustrated-html-dxpath}
\end{figure}

This annotation takes the form of a \textit{dynamicXpath}. A dynamicXpath is a (prefix, tag, suffix)-tuple. The prefix is the xpath leading up to a container HTML element. The tag is the root HTML tag of every child inside the parent container. The suffix is the xpath from the root child tag to the element inside the child which ought to be clicked. Figure~\ref{fig:illustrated-html-dxpath} shows how dynamicXpaths can be used to retrieve children containing the desired click targets from a page. Clicking on any of the identified target elements inside these children would resolve the click event at task execution time.

\subsubsection{Event Descriptions}

Next, OdoBot annotates the graph vertices corresponding to user-interaction events with general descriptions generated by summarizing the descriptions of all the events of the vertex. These event descriptions were generated previously in Section~\ref{sec:trajectory-collection}. The summarized event descriptions are added in an annotation property for the event (see the data entry and click events in Figure~\ref{fig:graph-model-sample}). Event-level descriptions allow OdoBot to produce natural language descriptions of paths through the graph by concatenating any annotations along the way.

\section{Model-driven Task Execution}
\label{sec:task-execution}

OdoBot's task execution involves three steps: (i) identifying the graph vertex that corresponds with the completion of the task; (ii) identifying the vertex that represents the current application state; and (iii) following a path between the two.

\paragraph{Identifying the Source and Destination Vertices} 
Given a natural language description of a task instance, OdoBot prompts its LLM to re-write this description as a general \textit{task}, omitting any specific parameters. This general task description is converted into a vector using OpenAI's text-embedding-3-large model, which is used to query the vector database of trajectory descriptions. The top-K (k=15) trajectories are returned, and an LLM is prompted to select the trajectory description that best aligns with the re-written task description. 
In the experimental runs, 65.7\% of correct trajectories are included with k@1, 90.4\% with k@5, 98.7\% with k@10 and 99.1\% with k@15. The chosen trajectory becomes the \textit{similar high-level task} used to guide path selection.

OdoBot then prompts its LLM to look through the API call and GraphQL mutation events associated with the similar high-level task and identify the one that best corresponds with the completion of the task. The graph vertex that contains the chosen event's ID is the destination vertex. Finally, OdoBot retrieves the current location of the browser (URL in the address bar) and identifies the graph vertex that contains location events with this URL. This becomes the source vertex.

\paragraph{Computing Paths from Source to Destination Vertices}
A modified beam search algorithm is used to plan paths from the source to the destination vertex. To that end, a metric is defined for estimating the distance between two vertices,
$v_s$ and $v_d$. 

OdoBot computes $T$, the set of trajectories that contain events mapped to both $v_s$ and $v_d$, and $T_e$, the set of events contained in all these trajectories. Next, the intersections of $T_e$ and $v_s$, $(T_e \cap v_s)$, and $T_e$ and $v_d$, $(T_e \cap v_d)$, are computed. 
For each of these two intersection sets, OdoBot computes the average index of their events. The absolute difference between these averages is the estimated distance between $v_s$ and $v_d$. 

\vspace{-0.4cm}
\begin{multline}
 dist(v_c,v_t) = \\|\textbf{avg}(\textbf{index}(e)|\forall  e \in (T_e \cap v_c)) - \textbf{avg}(\textbf{index}(e)| \forall e \in (T_e \cap v_t)) |
\end{multline}

The intuition underlying this distance metric is to use the average distance observed between events in their trajectories to estimate the distance between event vertices in the graph. If two vertices have no trajectories in common the distance between them is set to infinity. This represents OdoBot never having seen a trajectory linking those two vertices together. 

When exploring candidate paths via beam search, OdoBot initializes the set of paths being explored starting from the source vertex. Any vertex connected by a directed edge from the source vertex is a candidate vertex to explore. The distance metric above is computed for each candidate vertex and the top three vertices over this ranking are selected for the next iteration of the search. Paths with infinite distance steps continue to be explored until at most five infinite distance steps are included in a candidate path. This allows OdoBot to find paths which `bridge' gaps between observed trajectories, for example in situations where an unrelated trajectory demonstrates how to navigate to a vertex that does contain underlying events that share their trajectory with the target vertex.

\paragraph{Path Selection and Execution}
If no paths can be found between the source and destination vertex, OdoBot terminates. If a single path is found, that path is used automatically. If multiple paths are found, a path is selected based on two things: a natural language description of the path, and the \% of vertices that each path shares with the the similar high-level task. The natural language description of a path is constructed by concatenating the annotations associated with the events along the path (see Section~\ref{sec:model-annotation}). The IDs of the vertices containing events from the similar high-level path are compared with the IDs of the verticies in each path to produce an overlap \%. The path description and the overlap \% for all paths are included in a prompt to the LLM. The prompt instructs the LLM to select the path whose description best matches the natural language description of the given task instance, while preferring higher overlap \% paths. Since paths in the graph model can freely mix and match steps observed from different trajectories, path construction is step composition over steps learned from the trajectories. Preferring paths with higher overlap \% with the similar high-level task reduces the opportunity for composition errors that can arise when combining steps.

Once a path is selected, OdoBot initializes a cursor at the first vertex of the chosen path. Each vertex along the path is converted into a corresponding action to be applied to the underlying application as shown in table~\ref{tab:instruction-table}. OdoBot monitors its progress by capturing a live trajectory. When a new event appears in the live trajectory, OdoBot advances the cursor to the next event in the path only if the event maps to the event vertex the cursor is currently pointing to. This process repeats until the target vertex is reached, or a new trajectory event is not observed for longer than a predefined timeout period.

Application events are especially helpful in keeping OdoBot on track as a mismatched application event can indicate that something unexpected has happened. For example if OdoBot expects to be taken to the dashboard location after clicking the login button, but instead is taken to a course's discussion page, OdoBot will attempt to `re-locate' itself in the graph and compute a new path to the target vertex. The cursor is then reset to the first event in the recomputed path.

\begin{table}[h!]
    \centering
    \includegraphics[width=\linewidth]{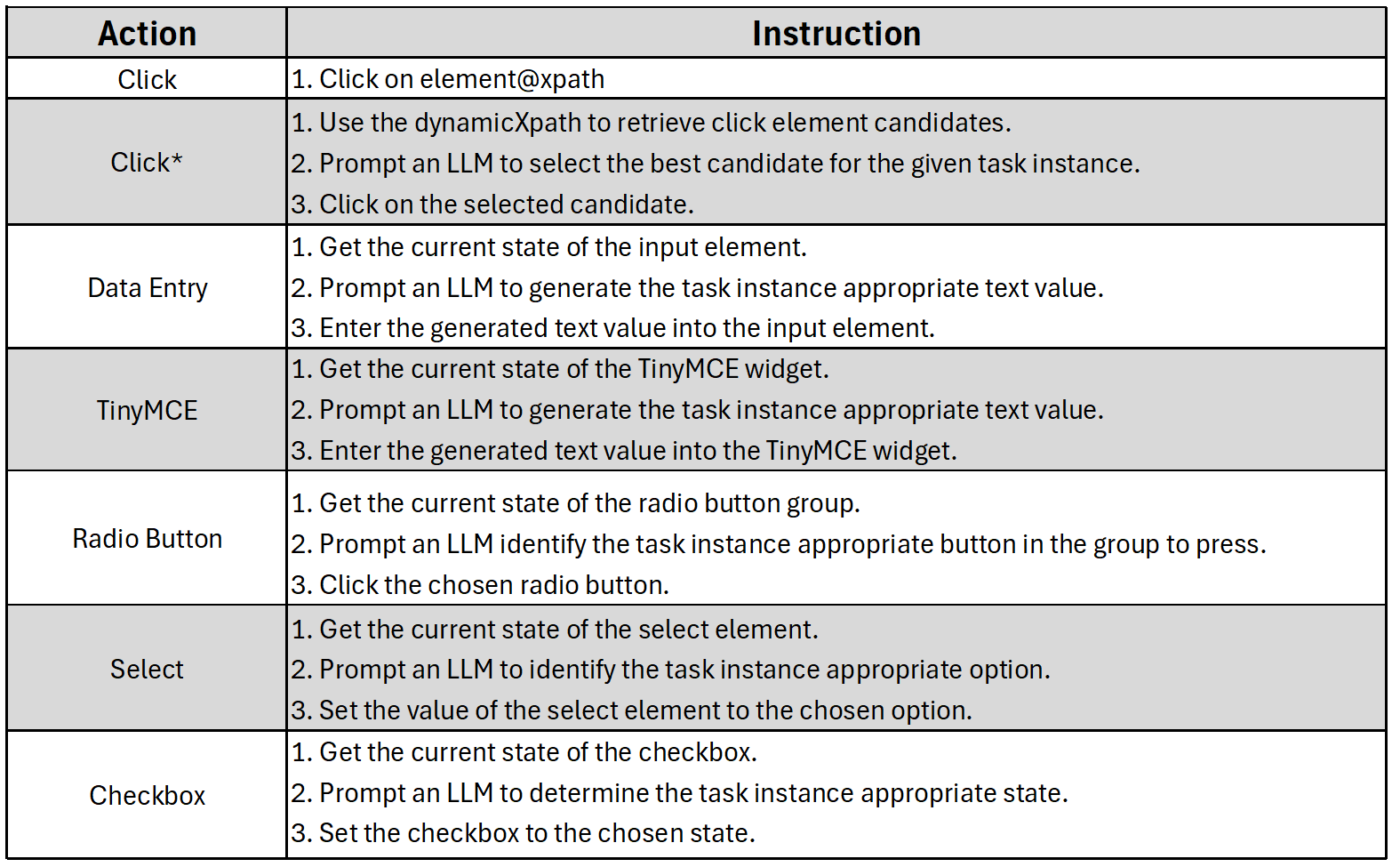}
    \caption{A table mapping actions in the graph to the instructions executed by OdoBot to resolve the action at task execution time. Click* refers to click actions annotated with dynamicXpaths.}
    \label{tab:instruction-table}
\end{table}
 
\section{The Experimental Sandbox}
\label{sec:sandbox}

We chose to evaluate our methodology with a real-world application, the Canvas Learning Management System (LMS). To create this Canvas-based evaluation environment, we had to construct a series of tasks and for each task we had to create all their relevant data in the environment. For example, a task involving a text submission for an assignment, requires the existence of a `student' who takes a `course' taught by an `instructor' who defined the `assignment' to which the student can submit some text. In this case, the task-relevant data consists of things like the names of the student, instructor, course, and assignment. 

Instead of hard-coding specific values for this data, we developed a script capable of generating all relevant Canvas entities, using mock data for a set of task templates. The script outputs a set of specific task instances that can be completed on the environment. For each task instance, the script also notes the API expected to be called upon the task completion and its parameters. 
For example, in the case of the above `submit a text entry for an assignment' task, the script will generate a course with ID `8' that has at least one assignment that allows text submissions and has an id `5'. The script will also report that, when this task is completed, a POST HTTP request will be invoked with the path `/courses/8/assignment/5/submissions' with some text in the body.

\paragraph{Task Selection} To produce our set of evaluation tasks we fed Canvas' 165 student guide pages~\cite{canvas-student-guide} to OpenAI's `gpt-4.1-2025-04-14' model, prompting it to produce a task description for each one of its manual pages. The model was instructed to create tasks with a single objective (i.e., no conditional or optional objectives); incorporate some or all of the features described on the page; and include all necessary inputs to complete the task; where references to specific courses/assignments/discussions/etc. were appropriate, the model was instructed to make up their names.

Next, each generated task description was presented to the same model to extract its inputs and their types. For example, if `Psychology 101' was referenced in the task description, the model was expected to identify `course' as one of the task inputs. 
This step generates a parameterized task template for each task instance emitted from the first step above, i.e., a template with labeled variables that can be substituted with different values to produce multiple instances of the same task. Listing~\ref{lst:parameterized-task-sample} shows an example of a parameterized task description. 

\begin{codelisting}
\vspace{.2cm}\hrule
\caption{\label{lst:parameterized-task-sample} A task template, generated from a page in the Canvas student guide, after manual curation. }
\begin{Verbatim}[breaklines=true]
Task: In your group "[[Group]]" for the course [[Course]], create a new announcement with the title "[[Announcement]]" and the following content: "[[Announcement Message]]" then publish it.
\end{Verbatim}
\hrule\vspace{.2cm}
\end{codelisting}

The resulting task templates were then manually inspected to identify side-effect producing tasks, i.e., tasks that cause changes to the application state through API calls and GraphQL operations. These tasks are the focus of our work. Thus, we excluded tasks involving interactions with third party plugins/tools/or features such as OneDrive, GoogleDrive, Zoom, etc.; separate physical devices, like webcams or printers; the to-do sidebar widget, the mobile version of the site, Canvas for Elementary; analytics or chat features; and any actions not associated with course content. As well, we excluded tasks downloading or uploading files and tasks simply seeking information based on specific content visible on the application pages. 45 side-effect tasks remained after the above exclusion criteria was applied. Each of them was manually verified and revised as necessary. Erroneous parameters, parameters not referring to Canvas entities were removed, ambiguous wording or instructions were clarified, and any step-by-step instructions copied from the student guide were removed.

\paragraph{Data Generation} Since all the entities required for the evaluation tasks were associated with courses, it was possible to manually create a YAML file containing mock data for a single course that contained mock data for all the necessary entities, such that all 45 tasks would be possible on that course. This YAML file was then used as a seed by a data generation script to create YAML files with the same structure but different data for N mock courses.

The script leveraged OpenAI's API prompting gpt-5-mini to produce novel realistic content for discussion entries, quizzes, assignments, etc. using the examples in the seed YAML to improve the generated output. Unwanted hallucinations were mitigated by generating complex entities one at a time, and automatically verifying generated data using known relationships between the data. For example, a fixed pool of student and instructor accounts was established at the start of the generation process, if a generated discussion reply later on listed an author that did not correspond with any previously established student or instructor accounts, that discussion reply was marked invalid and regenerated.

The data for ten mock courses was generated in this fashion. This course data was then used to construct a live, locally hosted Canvas environment. Since each course could facilitate the full 45 tasks, and data for 10 mock courses was generated, the resulting environment supports the execution of 450 unique task instances, 10 instances for each task template in the evaluation dataset. A snapshot of Canvas' database is then created to allow resetting and initialization of the environment directly to this state without the need to regenerate it.

The course and task data, evaluation scripts, as well as collected trajectories\footnote{https://anonymous.4open.science/r/CASCON-2026-OdoBot-Dataset}, and raw results\footnote{https://zenodo.org/records/21344539} are available.

\section{The Study}
\label{sec:study}

45 task instances, one from each of the 45 task templates, were selected for model construction using OdoBot.
The model construction and annotation for 45 trajectories consumed approximately 10.3 million tokens and was completed with OpenAI's `gpt-5.4-mini-2026-03-17' model. Trajectory recordings started from a clean environment loaded from the database snapshot but the environment is \textit{not} reset between each recording. Each trajectory begins from the Canvas login screen, proceeds until task completion, and terminates.

Once the OdoBot model was constructed, a different set of 45 task instances were selected for the comparative evaluation of OdoBot against WebVoyager and Agent-E. During task execution, the environment captures all network events and passes them as input to the evaluation script to determine if the task was completed successfully or not. Token use is also logged for each task. After all 45 tasks are completed, the database is reset, and the success rate is calculated for that run. Each agent completes the 45 tasks five times using gpt-5.5-2026-04-23 to establishing mean task success rates and token usage statistics. To measure how effective WebVoyager is at converting tokens into task success rate, three different configurations of WebVoyager are evaluated using screenshots of the size; 854x480 (480p), 1280x720 (720p) and 1920x1080 (1080p).

\section{Findings and Discussion}
\label{sec:findings}

\begin{table}[h]
\centering
\renewcommand{\arraystretch}{1.2} 
\begin{tabular}{|m{1.66cm}|m{1.35cm}|m{1.35cm}|m{2cm}|}
\hline

\textbf{Agent} & \textbf{Success Rate \unit{\%}} & \textbf{Total Tokens \unit{millions}} &\textbf{Tokens/Task} \unit{thousands}\\ \hline
OdoBot                     & $76.4 \pm 2.0$ & $0.9 \pm 0.02$ &$19.9 \pm 0.4$\\ \hline 
Agent-E                    & $85.3 \pm 2.5$ & $1.6 \pm 0.10$ &$35.2 \pm 2.3$\\ \hline  
WebVoyager {[}1920x1080{]} & $72.9 \pm 1.0$ & $4.6 \pm 0.07$ &$101.5 \pm 1.6$\\ \hline
WebVoyager {[}1280x720{]}  & $67.6 \pm 3.4$ & $3.2 \pm 0.09$ &$69.9 \pm 2.1$\\ \hline
WebVoyager {[}854x480{]} & $34.7 \pm 2.0$ & $2.5 \pm 0.05$ &$56.3 \pm 1.1$\\  \hline 
\end{tabular}
\renewcommand{\arraystretch}{1.0} %
\caption{Mean task success rate, total token use, and token use per task for each agent for the 45 evaluation tasks over five runs. 
}
\label{tab:task-performance-table}
\end{table}

\paragraph{Token Use}
OdoBot consumes 900 thousand tokens to achieve a task success rate of 76.4\% on our benchmark, outperforming WebVoyager's best (1080p) configuration in raw success rate. The superiority over WebVoyager is significant when evaluated with paired t-test (p-value 0.0161). Agent-E does achieve a higher (+9\%) raw task success rate, but requires 77\% more tokens per task to do so, a drawback even more pronounced for WebVoyager that requires 4.6 million tokens or 511\% more tokens. 
This reduction in token use arises from the fact that OdoBot never sends a full state observation to an LLM. Instead of delegating task execution logic to an LLM; task execution progress in OdoBot is monitored by comparing live trajectory events against expected events along the chosen execution path through the graph model, an LLM-free process, inline with externalization trends in this area of research~\cite{zhou2026externalizationllmagentsunified}.

When task-specific decisions must be made during a path's execution, OdoBot uses its library of common HTML sub-structures to produce an option selection prompt that only includes the task description, and the specific HTML elements to choose from, which are themselves filtered of all HTML attributes except value, type, option, placeholder, name, aria-label, id, action, alt, checked, for, form, href and title.

\paragraph{Error Analysis}
Agent-E fails 10 - 13 tasks; 3-4 tasks due to ethical refusal, 1-3 tasks because it fails to follow task instructions, 1-2 tasks due to failure to interact with UI widgets, and 1 task because it gets lost. WebVoyager fails 13 - 14 tasks; 8-9 tasks because it fails to interact with UI widgets; 1-2 tasks because it fails to follow task instructions, 1-2 tasks because it gets lost, and 0-1 tasks due to ethical refusal. OdoBot fails 10 - 12 tasks; 5 tasks fail due to modeling errors, 3 tasks fail due to failure to interact with UI widgets, 1 task because the required behavior was not captured in the recorded trajectories, 1 task because an xpath observed in the model could not be resolved in the live execution environment. The remaining 0-2 errors are stochastic and occur if the LLM makes incorrect choices at task execution time.

WebVoyager struggled to identify certain check boxes and radio groups, perhaps due to a lack of visual contrast for these elements. OdoBot struggled with tasks involving highly-dynamic content, especially if different elements replace others at the same xpath. This is common for  quizzes or surveys. For example question two, might be placed at the exact same xpath as question one, after clicking the `next question' button. The number of trajectories required to fully map out all possible states is high, and correct model mappings cannot rely so heavily on xpaths. 

At times, Agent-E and Webvoyager, got lost trying to find the group membership options. This UI is unintuitively located in the course section listing the groups, rather than the group settings. WebVoyager also got lost looking for instructor feedback on a quiz submission. OdoBot's model incorporated this information from past trajectories, and successfully completed these tasks in all its runs. Both WebVoyager and Agent-E attempted to refuse some quiz tasks, citing the ethics of completing a quiz on a user's behalf, albeit, in the one instance where WebVoyager refused, it had already entered all the answers, but refused to click the submit button. Since all agents were using the same LLM, the fact that the refusal rates varied between agents merits further investigation. OdoBot's lack of refusal is partially explained by its inability to complete most quiz tasks, nevertheless, OdoBot completed the quiz where all questions were presented at once in 4 of 5 runs, it is possible that new ethical safeguards have to be considered if LLMs become less involved in task execution.

\paragraph{Limitations}
While evaluating on a real-world application like the Canvas LMS offers insight into how web agents can perform outside of simulated sandbox environments, it also dramatically raises the effort required to construct meaningful tasks and the environment test data that allows them to be completable. The exclusions listed in the task selection section were necessary to constrain the effort required to create a Canvas environment in which the selected tasks could be verifiably executed by agents. 

OdoBot relies on xpaths to model events and execute actions, these are notoriously fragile~\cite{1771robula}. Decoupling xpaths from from the model construction process would attenuate this fragility. An interesting approach might be to extend the technique used to detect common HTML sub-structures, which produces `structural' fingerprints for every DOM element. Combined with an element-level screenshot it might be possible to create unique, reliable, non-xpath based identifiers for elements on the screen.    

At present trajectories are collected from human demonstrations of task executions, and are thus costly, but combining OdoBot's behavioral models with NetNav's unsupervised trajectory collection methodology would be a promising direction for future research~\cite{murty2025nnetnavunsupervisedlearningbrowser}.

Lastly, this work focuses on the cost in terms of token use since it represents the ongoing costs of operating such a system.

\section{Conclusion}
\label{sec:conclusions}
This work makes two contributions to the state of the art on web-agents research. First, it presents a new testbed based on widely used the Canvas LLM. This testbed enables web agents to interact with a live real application. The second, and most important contribution of this work is OdoBot, a web agent that exemplifies a novel web-agent methodology leveraging demonstrations of the web-application usage to construct a model of its behavior so that it can make more efficient use of LLMs during task execution. This methodology substantially reduces token use during task execution. OdoBot struggles in portions of task execution where the flexibility to deviate from past observations is important. On average it takes 15,257.80 fewer tokens than Agent-E to complete a task, dividing the 10.3 million token investment for model construction by the improvement gives 676 tasks to amortize the cost. For WebVoyager the cost is amortized over 120 tasks. OdoBot's token efficiency results from using observed behavior to intelligently reduce the amount of information needed to re-create the behavior in the future. A hybrid methodology, where execution control is passed between OdoBot-style and conventional web agent approaches could capture the best of both worlds. In a world were tokens are precious, it matters.

\bibliographystyle{IEEEtran}
\bibliography{references}
\end{document}